\pdfoutput=1

\documentclass[twocolumn]{wiley-article}
\usepackage{siunitx}
\usepackage[utf8]{inputenc}
\usepackage{graphicx} 
\usepackage{subfig}    
\usepackage{float}     
\usepackage{apalike}
\usepackage{setspace}
\usepackage{multirow}

\papertype{Original Article}
\paperfield{Journal Section}
 
\title{SWheg: A Wheel-Leg Transformable Robot With Minimalist Actuator Realization}
\author{Cunxi Dai}
\author{Xiaohan Liu}
\author{Jianxiang Zhou}
\author{Zhengtao Liu}
\author{Zhenzhong Jia}
\contrib{Cunxi Dai, Xiaohan Liu and Jianxiang Zhou contributed equally to this work.} 

\affil[1]{Department of Mechanical and Energy Engineering, Southern University of Science and Technology, Shenzhen, China}
\corraddress{Zhenzhong Jia, Department of Mechanical and Energy Engineering, Southern University of Science and Technology, 518000 Shenzhen, China.}
\corremail{jiazz@sustech.edu.cn}

\fundinginfo{Science, Technology and Innovation Commission of Shenzhen Municipality, Grant/Award Number: ZDSYS20200811143601004}

\runningauthor{DAI \scriptsize ET AL.}

\begin{document}

\begin{frontmatter}
\maketitle

\begin{abstract}
This article presents the design, implementation, and performance evaluation of SWheg, a novel modular wheel-leg transformable robot family with minimalist actuator realization. SWheg takes advantage of both wheeled and legged locomotion by seamlessly integrating them on a single platform. In contrast to other designs that use multiple actuators, SWheg uses only one actuator to drive the transformation of all the wheel-leg modules in sync. This means an $N$-legged SWheg robot requires only $N+1$ actuators, which can significantly reduce the cost and malfunction rate of the platform. The tendon-driven wheel-leg transformation mechanism based on a four-bar linkage can perform fast morphology transitions between wheels and legs. We validated the design principle with two SWheg robots with four and six wheel-leg modules separately, namely Quadrupedal SWheg and Hexapod SWheg. The design process, mechatronics infrastructure, and the gait behavioral development of both platforms were discussed. The performance of the robot was evaluated in various scenarios, including driving and turning in wheeled mode, step crossing, irregular terrain passing, and stair climbing in legged mode. The comparison between these two platforms was also discussed.

\end{abstract}
\end{frontmatter}

\section{Introduction}  \label{introduction}

\label{Intro}


Wheels and legs mobile robots are still the most popular form of ground locomotion platforms to negotiate two main terrain categories, namely flat and rough terrains~\cite{chen2017turboquad}. Generally, wheeled systems feature fast, smooth and power-efficient locomotion on flat terrain, but behave poorly on rough terrain, resulting in slippage or even immobility~\cite{kim2014wheel}. In contrast, legged robots excel at negotiating with rough terrain but can hardly compete with wheeled robots in terms of speed and efficiency on flat ground \cite{cao2022omniwheg}. Therefore, a hybrid wheel-leg transformable robot that enjoys the advantages of wheels on flat ground and employs the mobility of legs on rough terrain will likely have the best overall performance in various environmental conditions \cite{chen2014Quattroped,sun2017transformable}.

\begin{figure*}[htbp]
  \centering
  \includegraphics[width=5in]{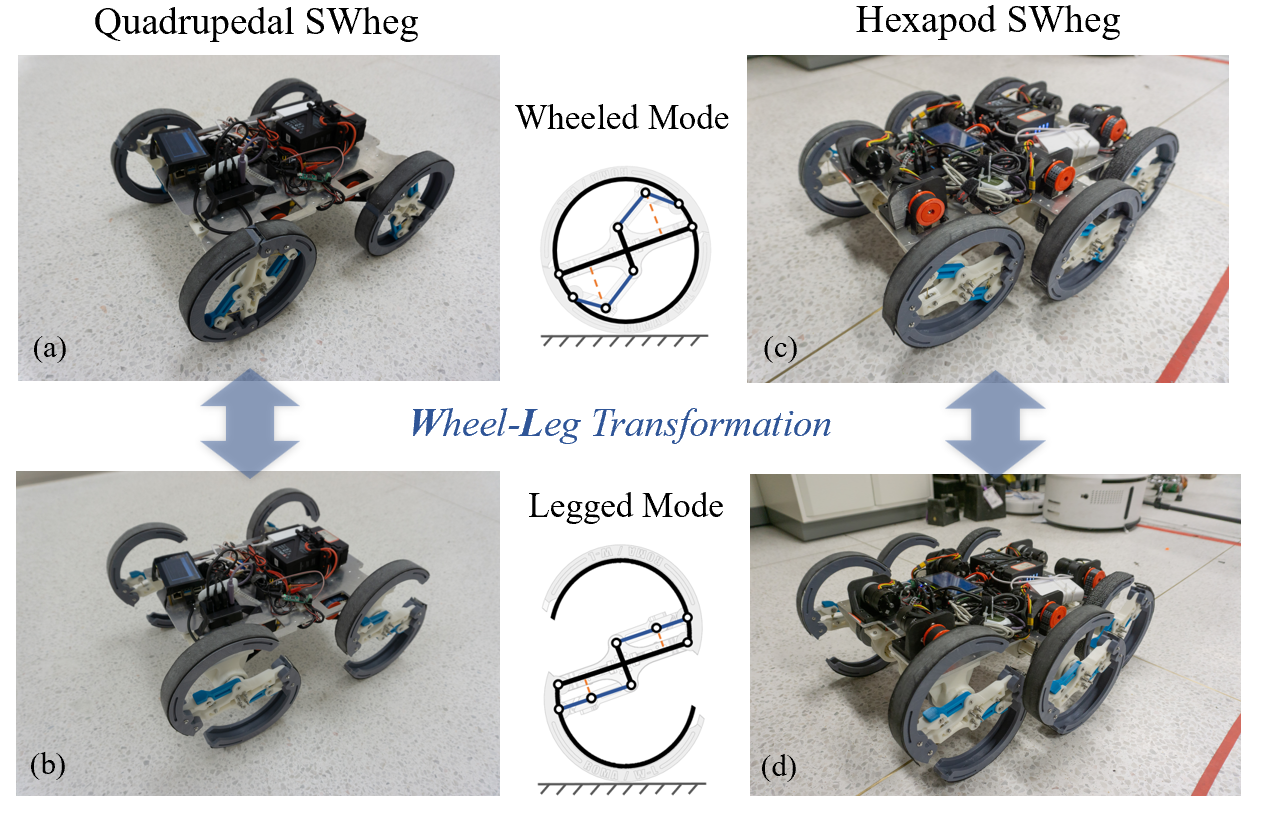}
  \vspace{-3mm}
  \caption{SWheg: Wheel-leg transformable robot with minimal actuator realization. (ab) Quadrupedal SWheg in wheeled mode and legged mode. (cd) Hexapod SWheg in wheeled mode and legged mode.}

  \label{fig:whole photo_intro}

\end{figure*}



Wheel-leg hybrid mobile platforms can be classified into three categories (details are given in the sequel) based on the hybrid strategy adopted, namely heteromorphic wheel robots, wheel-leg coexisting robots, and wheel-leg transformable robots.

Heteromorphic wheel mechanism usually features specialized fixed-shape wheels. For example, Loper has four $3$-spoke rimless wheels that perform well on both flat ground and stairs \cite{Bai2018An,Marques2006RAPOSA}. RHex and ION utilize six compliant half-circle "C-shapes" wheels, which are also referred to as wheeled-legged (WHEG) mechanism, to overcome obstacles. Benefited from its curved shape and compliant spoke \cite{Saranli2001RHex,agrawal2016ions}, it behaves reliably with only open-loop control with lesser torque required compared to a straight stiff leg \cite{Lee2017Origami}.

Another type of robot has a separate wheel and leg mechanism coexisting on the platform, and the locomotion is generated by the collaboration of these two mechanisms. For instance, Chariot III has two big wheels and four $3$-DOF legs \cite{Nakajima2004Motion}. PEOPLER-II has two bars mounted on each of the four wheels, and locomotion can be switched between leg type and wheel type \cite{Kosugi1984Motion}.Wheeleg has two pneumatically actuated $3$-DOF front legs and two independently driven rear wheels \cite{Lacagnina2003Kinematics}.

Different from the above robots, wheel-leg transformable robots feature transforming mechanisms that alter the morphology between wheels and legs. Some platforms utilize a passive $1$-DOF mechanism  that only transforms when encountering with a vertical plane of the obstacle, such as wheel transformer, land devil ray, and shape-morphing wheel \cite{6631385,Bai2018An,8979177}. Some platforms use an active $1$-DoF mechanism, and one of them uses origami to transform its wheel \cite{Lee2017Origami}. A robotic wheel overcomes an obstacle by pushing the spoke that forms a circular wheel \cite{Moriya2020Robotic}. FUHAR opens its spoke, which is referred to as a finger, when it climbs stairs \cite{R2020FUHAR}. The $1$-DOF mechanism can only be used to overcome a small range of obstacle sizes. Therefore, to overcome this limitation, $2$-DOF mechanisms are have been investigated. The Quattroped and STEP are examples of platforms using the $2$-DOF mechanism \cite{chen2013quattroped,RN91}. The Quattroped changes its wheel into the WHEG mechanism and move the wheel up and down. This transformation is used to set a strategy to climb obstacles.


Many of the wheel-leg hybrid robots mentioned above select "search and rescue" as their main tasks, such as \cite{campbell2003stair,moore2002reliable}. These working environments often have diverse terrains, requiring robots to have high-speed mobility over different types of terrains in order to complete the tasks in a timely manner because of the urgent nature of these tasks. 
Hence, wheel-leg transformable robots that can switch to suitable motion modes (i.e., wheeled mode on flat ground, legged mode in rough terrain) can be very competitive in such tasks~\cite{tadakuma2010mechanical}. However, the great overall performance of the wheel-leg transformable robot comes with a price, i.e., system complexity. 
They often feature more complicated mechatronics systems compared to robots with a single locomotion modality. Most importantly, the extra DoFs in the transformation mechanism calls for more actuators, which increases the systems' cost and uncertainty dramatically.
For example, the hexapod Slegs robot reported in ~\cite{soyguder2017slegs} has at least $18$ actuators, i.e., $3$ actuators for each wheel-leg module. Slegs robot uses $6\times1$ main actuators to drive the $6$ wheel-leg modules and additional $6\times2$ actuators (servos or linear actuators) to switch between the wheeled and legged modes. The Quattroped robot in \cite{chen2013quattroped} and the turboquad robot in \cite{chen2017turboquad} are both quadrupedal robots with at least $8$ actuators. The turboquad robot uses $4\times2$ main actuators to drive the specially-designed wheel-leg module, which can complete mode switching online. It also uses an additional actuator to control the Ackermann steering mechanism. Hence, it has $9$ actuators in total. 
To our best knowledge, all wheel-leg transformable robots reported in the literature with $N$ wheel-leg modules will have at least $2N$ actuators. However, the independent transformation control of each wheel-leg module is not a key function to the robot, Theoretically, one additional actuator is sufficient to actuate the transformation of all the wheel-leg modules. 



In this paper, we proposed a tendon-driven wheel-leg transformable robot design with minimal actuator realization, as shown in Figure~\ref{fig:whole photo_intro}. In particular, for a robot with $N$ wheel-leg modules, we only need $N+1$ actuators. This is only about $50\%$ of other literature designs. This will help reduce the system complexity, cost, failure rate, and the transformation of all the modules are synchronized by root.

The main contributions are:

\begin{enumerate}
  \item We proposed the design principle of a tendon-driven wheel-leg transformable robot with minimal actuator realization. In particular, the $N$-legged robot only needs $N+1$ actuators, with only one actuator for transformation. This can greatly reduce the system's complicity and cost.
  \item We designed a novel wheel-leg transformable module that used a tendon-driven mechanism for locomotion mode switching. We also proposed a design that can use a single actuator to control all tendons to finish mode switching of the entire robot.
  \item We proposed the design and performance evaluation of a new S-shape leg morphology and built two robot platforms to validate our design, quadrupedal SWheg, and hexapod SWheg. We also developed control algorithms for the two platforms and validated the effectiveness in both simulation and field experiments.

\end{enumerate}

 The remainder of this paper is organized as follows. Section~\ref{transformable wheel} presents the design of the transformable wheel mechanism and the $4$-bar linkage mechanism to create the transformation. Section~\ref{Robot Platform Design} presents the tendon network and the mechatronics infrastructure of the platform. In Section~\ref{controller design}, the design and performance of the modular wheel are validated. We also performed comparative power experiments on quadrupedal and hexapod robots. The results are discussed in Section~\ref{experiments}.

\section{Transformable Wheel Mechanism} \label{transformable wheel}
\label{gen_inst}

\begin{figure*}[t!]
    \centering
    \includegraphics[width=5.5in]{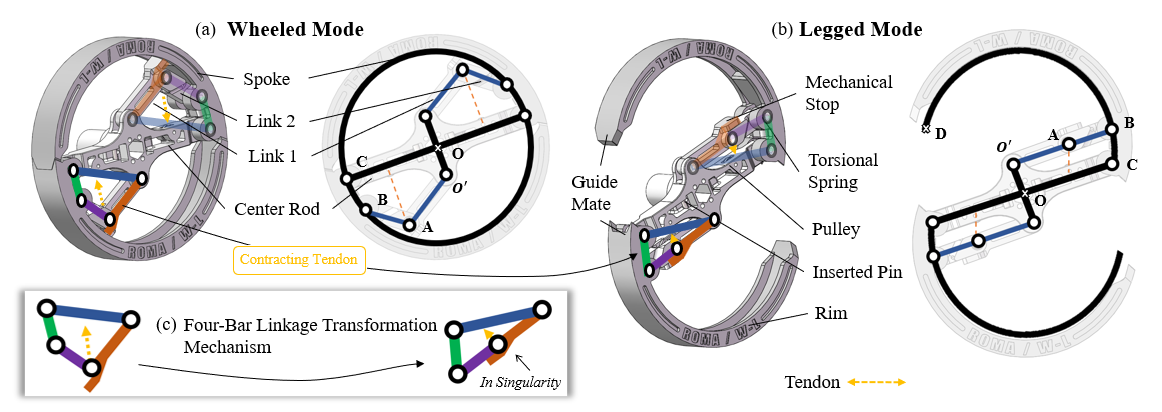}
    \vspace{-3mm}
    \caption{The mechanical design of SWheg's Wheel-leg transformable module. (a) and (b) are the mechanical structure of the wheel-leg transformable module, the parallel four-bar linkage transformable mechanisms are labeled. The notations for the links and joints used in this paper are also marked. (c) The schematic diagram of the transformable mechanism. The mechanical stop is drawn as part of link $1$. The tendon are shown in dashed line, the arrows indicates the direction of the force exerted by the tendon.}
    \label{fig:tendon network}
    \centering
\end{figure*}

\begin{figure}[t!]
    \centering
    \includegraphics[width=2in]{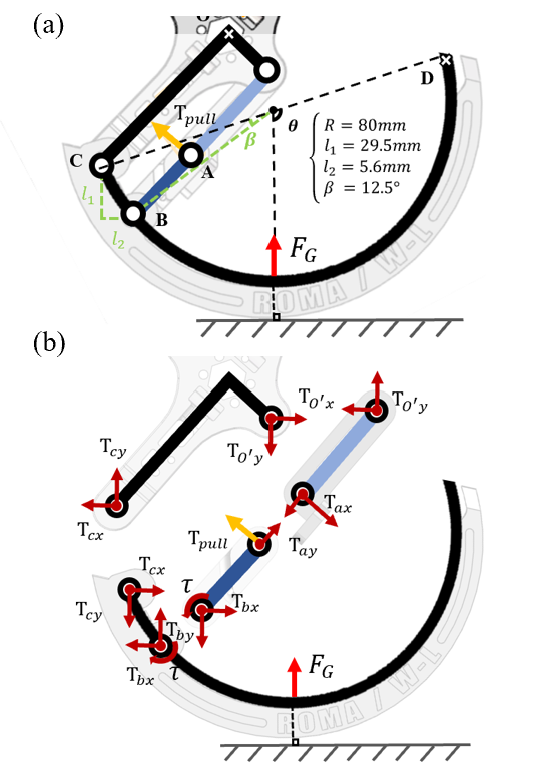}
    \vspace{-3mm}
    \caption{Free body analysis of the wheel-leg module in legged mode. (a) Parameters of the S-shaped leg mechanism. (b) Notations used in Free body analysis.}
    \label{fig:WheelModeSch}
    \centering
\end{figure}

\begin{figure*}[t!]
    \centering
    \includegraphics[width=5.5in]{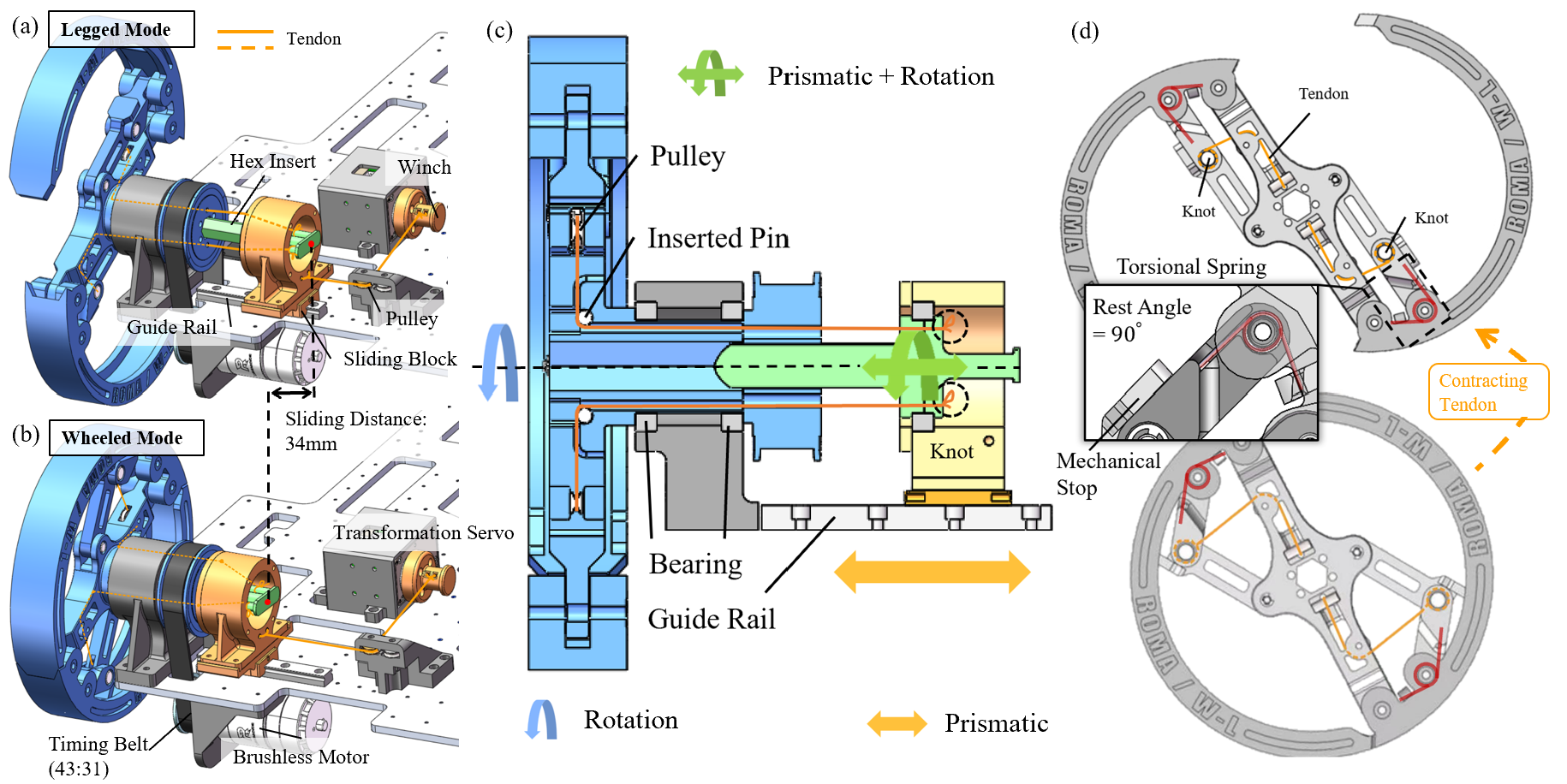}
    \vspace{-3mm}
    \caption{Detailed mechanism design of SWheg's wheel-leg transformable module. (a) shows the part status of the module in legged mode, and (b) shows status in wheeled mode. When transforming from (b) wheeled mode to (a) legged mode, the transformation servo contracts the tendon and pulls back the 2-DoF rotating connector by 34mm. The Hex insert slides with the connector and rotate passively with the wheel to prevent winding. (c) The section view of the wheel-leg transformation mechanism is shown on the left side, showing the embedded tendon arrangement. (d) shows the placement of the inserted torsional spring and tendon connection during transformation.}
    \label{fig:WheelModeSch}
    \centering
\end{figure*}

In Legged robots, the geometrical configuration of the leg strongly affect the legged motion agility and energy performance \cite{chen2014Quattroped}. Among these robots, "C-shape legs" or "half-circle legs" with curved shape and compliant spoke were a popular design choice \cite{Saranli2001RHex,agrawal2016ions,chen2014Quattroped}. This leg geometry has been proved to behave reliably with only open-loop control compared to a straight stiff leg \cite{Lee2017Origami}. 
"S-shaped legs" have two half-circle rims on a single leg, and has been previously studied in wheel-leg transformable robots such as TurboQuad and SLegs. It features similar curvature and compliant area as "C-shaped legs", and has also been proved to generate smooth legged locomotion. Moreover, ”S-shaped legs“ are more geometrically friendly to design a wheel-leg transformation mechanism, because it has two half-rims instead of a single half-rim in "C-shape legs" or "half-circle legs". Therefore, we decide to use a ”S-shaped legs“ to be the legged geometry for SWhegs \cite{chen2017turboquad,soyguder2017slegs}.

One of the key challenges to the wheel-leg transformation mechanism is to sustain legged mode without over-demanding the central servo or consuming too much power. Both of these tasks can be solved if the mechanism reaches singularity in legged mode. Towards this goal, the transformation mechanism was designed based on a four-bar parallel mechanism that reaches singularity in legged mode, so that minimum force is required from the center servo. Two four-bar linkage mechanism was placed on a single wheel-leg transformation module as marked on Figure~\ref{fig:tendon network}. Another benefit of this configuration in legged is that the spoke is supported at two points with a stable triangle structure formed by the transformation mechanism in singularity.

The wheel-leg transformation mechanism contains two identical four-bar parallel mechanism, each support one spoke to expand from the center rod. When fully expanded, the two spokes form a "S-shape" wheel that is compliant in most area. An additional mechanical stop is designed to ensure that the mechanism reaches singularity when fully expanded. The schematic diagram of the transformation mechanism is shown in Figure~\ref{fig:tendon network}. To optimize the transformation mechanism design, the following kinematic and requirements and goals should be set.
The geometric model of the wheel is set up using MATLAB, and the values of four-bar mechanism design variables are searched by trial and error to satisfy the previous kinematic requirements. The workspace and the singularity of the transformation mechanism is shown in Figure~\ref{fig:tendon network}.

\subsection{Mechanical Design of the Modular Wheel}
The wheel-leg transformable module is mechanically self-contained except for the transformation driving tendon, which is driven by central servo. The tendon arrangement is marked on the bottom view of the quadrupedal and hexapod platform in Figure~\ref{fig:tendon network}. 
Since the tendon can only exert force in a single direction, restoring from legged mode to wheeled mode requires an additional source of force or torque. We experimented with multiple restoring methods, such as rubber bands and linear springs. The common problem with these linear elastic elements is that it does not provide powerful enough force to maintain wheeled mode.
So we decided to use a torsional spring to provide restoring torque on the joint directly.

The torsional spring is embedded concentrically with joint $B$ to provide the torque to restore from wheeled mode to legged mode. The trade-off on selecting the spring stiffness is that: if the spring is too soft, it cannot maintain wheeled mode. If the spring is too stiff, the central servo will exert extra unnecessary torque to drive the transformation mechanism. So we tested with various stiffness and landed on a spring with spring constant $k=0.43Nm/rad$, which is the minimum stiffness that can maintain the wheeled mode. The stiffness is determined
When transforming from wheeled mode into legged mode, the central servo rotates the winch and pulls the tendon to drive the transformation mechanism on each wheel-leg module. A mechanical stop is added to link$1$ so the transformation mechanism does not cross singularity. When restoring from legged mode to wheeled mode, the central servo rotates reversely and release the tension on the tendon, the torsional spring provides the required torque to restore to wheeled mode.

\begin{figure}[t!]
    \centering
    \includegraphics[width=2in]{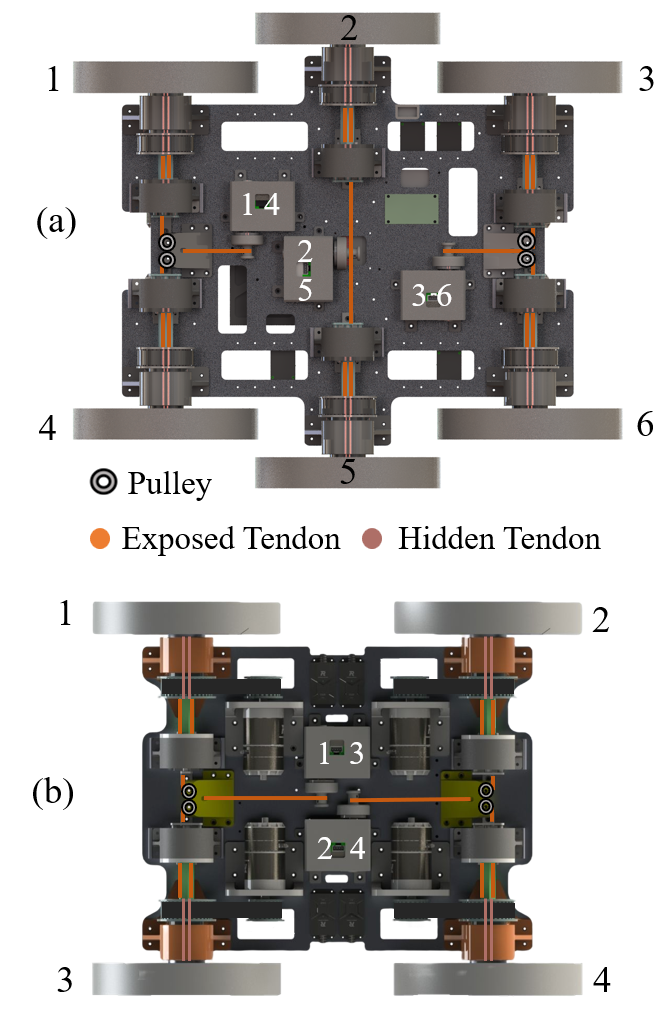}
    \vspace{-3mm}
    \caption{Tendon network of the two SWheg robots, (a) hexapod SWheg and (b) quadrupedal SWheg. The numbers shows the grouping between servos and wheel-leg modules.}
    \label{fig:WheelModeSch}
    \centering
\end{figure}

An inevitable problem with the tendon driven actuation method is that the tendon rotates with the wheel while the driving servo is fixed. This would result in entangled tendon and must be solved. The most direct way to approach this problem is to separate the rotation and prismatic motion. To realize this concept, we designed a $2$-DoF connector in series with the tendon. The connector has one rotational DoF, one linear DoF and is connected in series between the central servo and the wheel-leg module, concentric with the wheel. The prismatic range of motion is limited by guide rail length and a mechanical stop. The details of this mechanism is shown in Figure~\ref{fig:tendon network}. The tendon from the servo winch pulls on the stator and pulls the connector slides on the guide rail, the rotor rotates passively with the wheel via the hex rod insert into the wheel-leg module.
There is only one motor in each module to drive the wheel, a belt transmission system comprising two pulleys and a timing belt with a $43:31$ speed reduction is used.

\subsection{Design Considerations Based on Quasi-Statics}
It is important to calculate the torques required for the mobile
central servo with transformable wheels to design. Without loss of generality, it is assumed that the weight of the platform is $15kg$ and each wheel supports a load of $5kg$. Also, it is supposed to move at a sufficiently low speed so that no slip occurs.
The free body diagrams (FBDs) of transformable wheel are shown in Figure~\ref{fig:tendon network}a. In singularity configuration, linkage $AB$ and the tendon only have to counter with the restoring torque provided by the inserted torsional spring, the following relations can be derived:
\begin{equation}
T_{pull} l_{AB} =\tau 
\end{equation}

Therefore, the required torque on the servo to sustain legged mode is a constant of 
$3.1N/m$ , directly related to the spring constant and irrelevant of the direction and amplitude of $F_{G}$, the ground reaction force.

\begin{enumerate}
    \item The mechanism forms a circle with $12cm$ radius when closed.
    \item The wheel-leg module has the largest compliant area $\stackrel\frown{BD}$ in the legged mode. 
    \item The mechanism should expand as much as possible, meaning to optimize the distance between joint $O$ and $D$ when fully expanded
    \item The four-link mechanism reaches singularity when fully expanded.
    \item Interference between the components during deformation must be avoided. 
\end{enumerate}
\section{Robot Platform Design} \label{Robot Platform Design}

To validate the ubiquitous feasibility of this minimal actuator realization design, and to investigate the influence of leg number on a robot's performance, we built two mobile experiment platforms with four and six wheel-leg transformation modules, namely quadrupedal SWheg and hexapod SWheg as shown in Figure~\ref{fig:whole photo_intro}. 
We used two servos on the quadrupedal SWheg to drive the transformation mechanism of the front wheels and back wheels separately, and correspondingly three servos on the hexapod SWheg. Doing so enables the robots to transform each pair of wheels individually thus making it easier to investigate the performance of a quasi wheel-leg robot.
However, it has been proved in Section~\ref{transformable wheel} that it is feasible to use only one servo to drive the overall transformation.

\subsection{Tendon Network Design}
The tendon network on the robot platform is constructed in a way that allows a single actuator to control all tendons for mode switching of the entire robot.
The transformation servo group is placed in the geometrical center of orthographic projection of the platform. Pulleys are placed to ensure the tendon pulls on the wheel-leg transformation module vertically.
The transformation tendon is knotted at joint $A$ This ensures that the tendon is contained in the envelope of the wheel in both wheeled and legged mode, and is therefore not exposed to unexpected environment and risk of being cut. The tendon goes through the center rod and connects with the connector mechanism. Pins and pulleys were embedded in the transmission route to reduce transmission friction on the rod.

\subsection{Mechatronics Infrastructure}
\begin{figure*}[t!]
    \centering
    \includegraphics[width=5in]{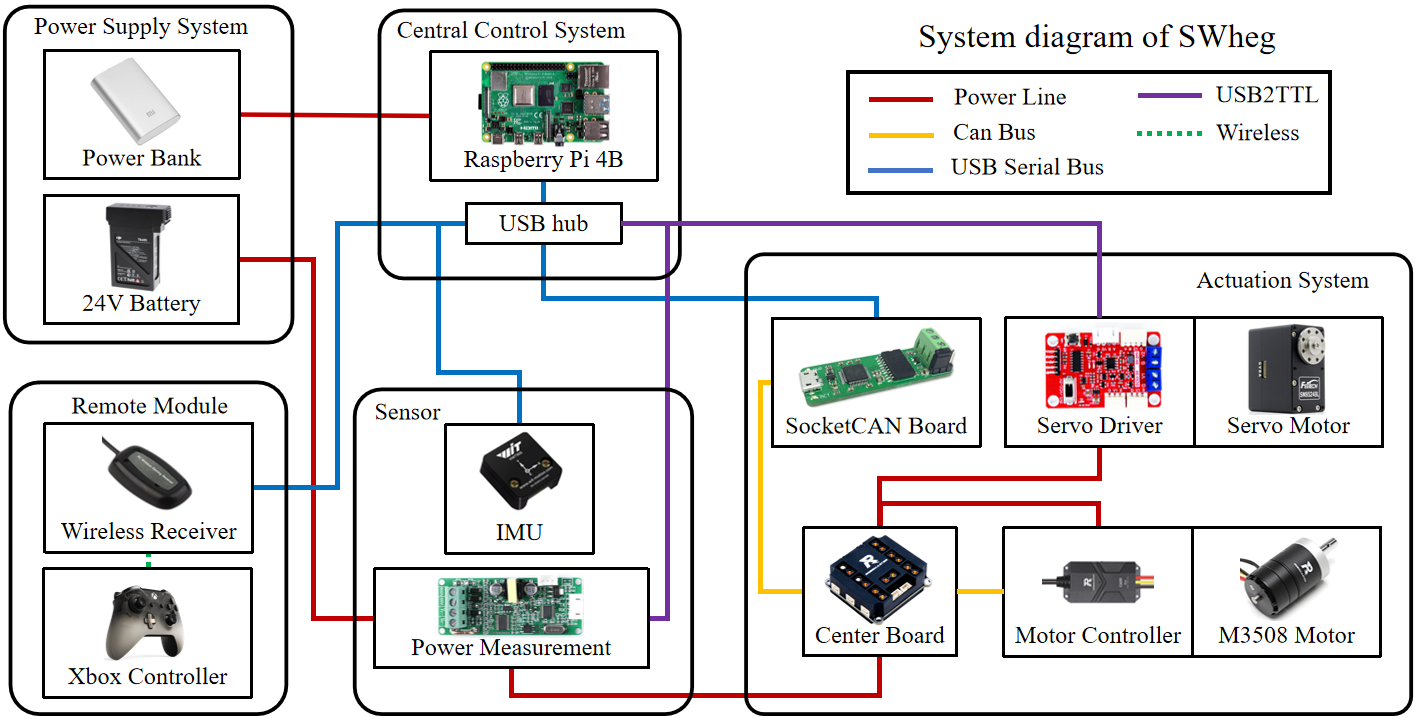}
    \vspace{-3mm}
    \caption{System diagram of SWheg.}
    \label{fig:system_diagram}
\end{figure*}

As shown in Figure~\ref{fig:system_diagram}, the main computation power on this robot is the Raspberry Pi 4B embedded computer, and Robot Operating System (ROS) is used on this board for inter-module communication. And we use the Xbox controller as a remote control module. The IMU and power measurement sensor are installed on the chassis for measuring motion smoothness and power consumption. About the actuator system, take the quadrupedal SWheg for an example, $4$ DJI M3508 brushless geared motors are used as the major power unit for this robot. Each motor is connected to a DJI C620 FOC controller. The controllers accept the current command and provide current, speed, and position feedback. A socketcan board that converts the CAN signal to the USB signal is used to connect the FOCs and the embedded computer. Two Feetech SMS serial servos are used to power the tendon-driven system. The servos are connected to a servo driver board which provides power and signal exchange. The servo driver board is also connected to the embedded computer. The specifications of the robots are summarized in Table~\ref{tab:ROBOT SPECIFICATIONS}.

\begin{table*}[]
\caption{ROBOT SPECIFICATIONS}
\label{tab:ROBOT SPECIFICATIONS}
\begin{tabular}{lccc}
\hline
\multicolumn{1}{c}{}      &                                & Quadrupedal SWheg            & Hexapod SWheg                \\ \hline
\multirow{2}{*}{Length}   & Body                           & 0.4m                         & 0.47m                        \\
                          & Hip-to-hip                     & 0.3m                         & 0.2m                         \\ \hline
\multirow{2}{*}{Width}    & Body                           & 0.3m                         & 0.3m                         \\
                          & Leg-to-leg                     & 0.39m                        & 0.46m                        \\ \hline
\multirow{3}{*}{Height}   & Body                           & 0.154m                       & 0.154m                       \\
                          & Ground to hip (legged mode)    & 0.0965m                      & 0.0965m                      \\
                          & Ground clearance (legged mode) & 0.193m                       & 0.193m                       \\ \hline
\multicolumn{2}{l}{Wheel-leg (i.e.,   rim) diameter}       & 0.0965m                      & 0.0965m                      \\
\multicolumn{2}{l}{Maximum radius of   wheel-leg}          & 0.1250m                      & 0.1250m                      \\ \hline
\multirow{4}{*}{Weight}   & Total                          & 8.406kg                      & 9.791kg                      \\
                          & Body                           & 6.392kg                      & 7.12kg                       \\
                          & Wheel-leg(each)                & 0.336kg                      & 0.336kg                      \\
                          & Battery                        & 0.673kg                      & 0.673kg                      \\ \hline
\multirow{2}{*}{Actuator} & Driving                        & DJI M3508 motor(×4)          & DJI M3508 motor(×6)          \\
                          & Switch mechanism               & Feetech SMS serial servo(×2) & Feetech SMS serial servo(×3) \\ \hline
\multirow{2}{*}{Sensors}  & IMU                            & (×1)                         & (×1)                         \\
                          & Power measurement              & (×1)                         & (×1)                         \\ \hline
\multirow{2}{*}{Battery}  & DJI TB48                       & (×1)                         & (×1)                         \\
                          & Power bank                     & (×1)                         & (×1)                         \\ \hline
\end{tabular}
\end{table*}
\section{Controller design} \label{controller design}
\label{others}
                    
\subsection{Control System Overview}

\begin{figure}[t!]
    \centering
    \includegraphics[width=2.5in]{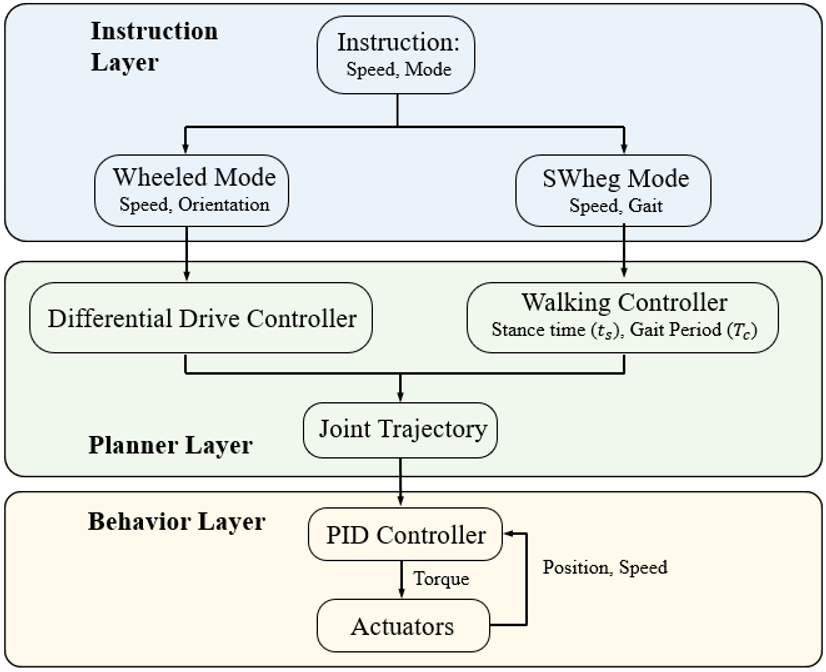}
    \vspace{-3mm}
    \caption{Structure of SWheg controller.}
    \label{fig:ControllerHierarchy}
    \centering
\end{figure}

The control system consists of three layers: the instruction layer, the planner layer, and the behavior layer. The instruction layer processes high-level instructions, including control mode, walking gait, and moving speed. A finite state machine manages controllers of different modes. During mode transitions, the joints' position and control mode will be adjusted  accordingly. For now, the instructions are provided by a human operator. The desired speed, gait, and velocity are then sent to the controller of the corresponding mode. In the planner layer, the controller generates velocity/position trajectories for each joint according to the command. The behavior layer controls the actuators. The actuators will track the provided trajectory with PD controllers.

\subsection{Wheeled mode}
When the robot operates in wheeled mode, the control of the joints is straightforward. The rotation velocity ($\omega_{i}$) of each joint is assigned based on the differential-wheel steering model. Joints on the same side of the robot will be assigned the same angular velocity, $\omega_{Left}$ or $\omega_{Right}$.
As shown in Figure~\ref{fig:WheelModeSch} when the robot has a forward velocity $v_{desired}$ and a rotational velocity $\omega_{desired}$, its instant center of rotation will be located at point $P_{ICR}$. The turning radius can be expressed as:

\begin{equation}
R_{c} = \frac {v_{desired}} {\omega_{desired}}
\end{equation}
Then, the angular velocity of the joint on each side can be computed from
\begin{equation}
\omega_{Right} = \frac{ v_{desired} + \frac{\omega_{desired} \cdot d}{2}} {R_{wheel} N_{t}}\\
\end{equation}
\begin{equation}
\omega_{Left} = \frac{ v_{desired} - \frac{\omega_{desired} \cdot d}{2}} {R_{wheel} N_{t}} \\
\end{equation}

Where $d$ is the width of the chassis, and $N_t$ is the speed reduction ratio. The motions are shown in Figure~\ref{fig:LegModeSch}.  The wheeled mode control strategy can also be used in robots with more Wheel-legs.

\begin{figure}[t!]
    \centering
    \includegraphics[width=2.5in]{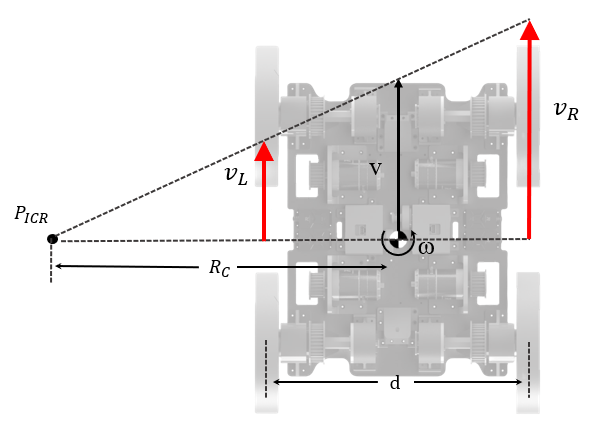}
    \vspace{-3mm}
    \caption{Differential-wheel steering model when operated in wheeled mode.}
    \label{fig:WheelModeSch}
    \centering
\end{figure}

\subsection{Legged mode}
In legged mode, the planner-layer controller will generate a group of clock-driven periodic trajectories for joints. This control strategy is joint space close-loop but task space open loop. The trajectories generated are periodic functions of time, which have outputs ranging from $0$ to $\pi$.

\subsubsection{Walking and Trotting}

\begin{figure*}[t!]
    \centering
    \includegraphics[width=5.5in]{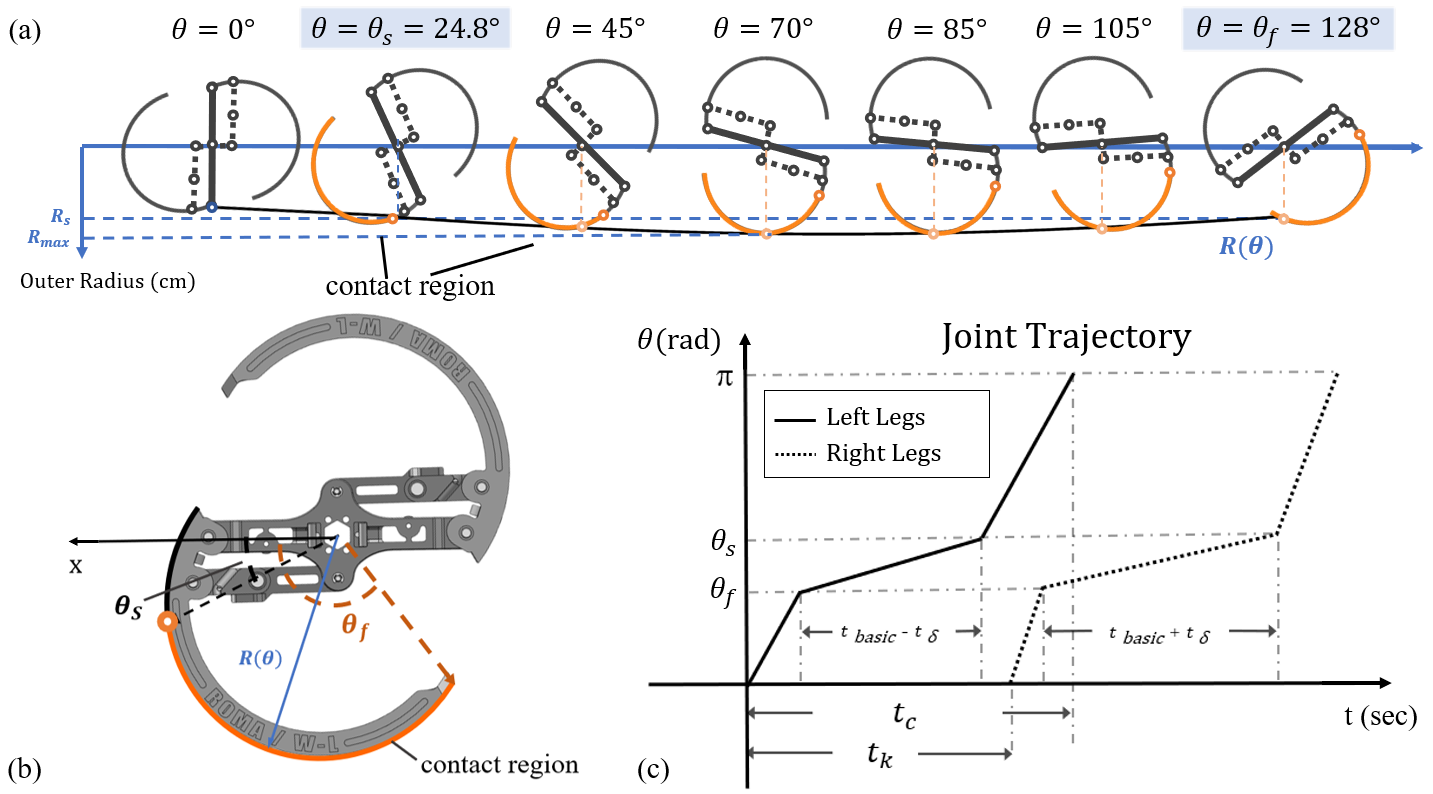}
    \vspace{-3mm}
    \caption{Schematic of leg mode. (a) Polar coordinate definition of the Wheel-leg. The polar axis x-axis of the polar coordinate system is set colinear with the center line of the center rod. (b) Joint trajectory in walking gait. (c) The outer radius of Wheel-leg's rim in legged mode over polar angle $\theta$. The approximated contact region is highlighted.}
    \label{fig:LegModeSch}
    \centering
\end{figure*}

For walking and trotting, a simple Rhex-like control strategy is implemented. As shown in Figure~\ref{fig:LegModeSch} (a) and (c), the wheel-leg has a larger outer radius for $\theta_{s} < \theta < \theta_{f}$ (the highlighted part). We suppose that this part of rim is the contact region which will touch the ground and the rest of the rim won't. In a cycle, a wheel-leg goes through a slow and a fast swing phase. A wheel-leg enters the slow phase when $\theta = \theta_{s}$, and enters the fast phase when $\theta = \theta_{f}$. The slow and fast phases are considered the stance and flight phases, respectively.

The Joint trajectory is shown in Figure~\ref{fig:LegModeSch}(c). The joint trajectory of walking mode has two key parameters: $T_c$, $t_s$, which determine the ground speed of a wheel-leg. $T_c$ is the period of the function, and $t_s$ is the approximate stance time. Then, the average ground speed at which a wheel-leg moves the body during its stance phase is approximately:

\begin{equation}
V_{leg} = \frac {R_{avg}(\theta_{f} - \theta_{s})}{t_{s}} 
\end{equation}

All six legs have the same period, $T_{c}$. The stance time of each wheel-leg can be expressed as  $t_s = t_{basic} + t_{\delta}$. All wheel-legs have the same $t_{basic}$. wheel-legs on the same side of the robot have the same $t_{\delta}$. Both $t_{basic}$ and $t_{\delta}$ are proportional to the period.  Then, the average ground speed $V_{side}$ provided by wheel-legs on one side can be expressed as:

\begin{equation}
V_{side} = \frac {R_{avg}(\theta_{f} - \theta_{s})}{k_{basic}*T_{c} + k_{\delta}*T_{c}} 
\end{equation}
Apply the differential wheeled model and we can have:
\begin{equation}
V_{forward} = \frac{1}{2} (\frac {R_{avg}(\theta_{f} - \theta_{s})}{k_{basic}*T_{c} + k_{\delta}*T_{c}} + \frac {R_{avg}(\theta_{f} - \theta_{s})}{k_{basic}*T_{c} - k_{\delta}*T_{c}} )
\end{equation}
\begin{equation}
V_{\omega} = \frac{1}{R_{car}} (\frac {R_{avg}(\theta_{f} - \theta_{s})}{k_{basic}*T_{c} + k_{\delta}*T_{c}} - \frac {R_{avg}(\theta_{f} - \theta_{s})}{k_{basic}*T_{c} - k_{\delta}*T_{c}})
\end{equation}
The forward linear velocity $V_{forward}$ is mainly determined by $T_{c}$. $k_{basic}$ is a constant. To ensure the robot's stability, $k_{basic}$ should be carefully chosen. From experiments, we found that $t_{basic} = 0.65 * t_{c}$ has a good performance on both quadrupedal and hexapod setup. By changing the $k_{\delta}$ we can roughly control the angular speed. Note that to keep the robot stable, the total stance time $t_{basic}+t_{\delta}$ has a lower limit of $\frac{1}{2} t_{c} $.

\begin{figure*}[t!]
    \centering
    \includegraphics[width=5.5in]{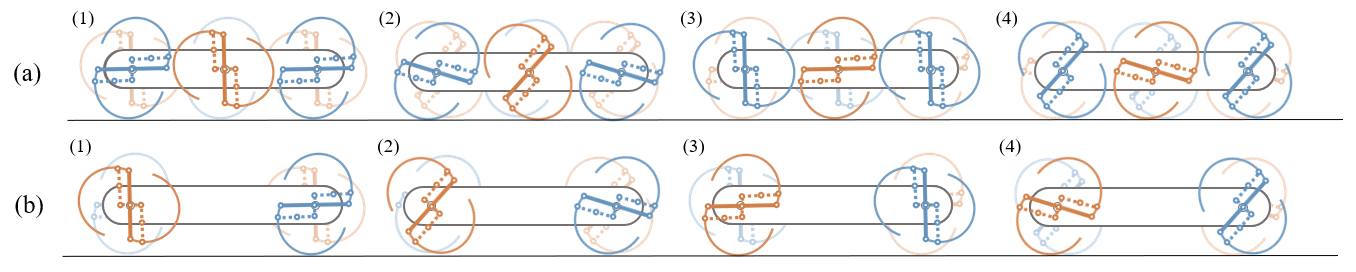}
    \vspace{-3mm}
    \caption{Schematic of (a) hexapod SWheg tripod gait and (b) quadrupedal SWheg trot gait. (1) of (a) and (b) : Initial state. (2) and (3) of (a) : Leg LF, LB, RM hit and then leave the ground. (2) and (3) of (b) : Leg LF, RB hit and then leave the ground. (4) of (a) : Leg RF, RB, LM hits the ground again. (4) of (b) : Leg LF, RB hits the ground again. The robot will then return to state (1). }
    \label{fig:gaitSch}
    \centering
\end{figure*}

\begin{figure}[t!]
    \centering
    \includegraphics[width=2.5in]{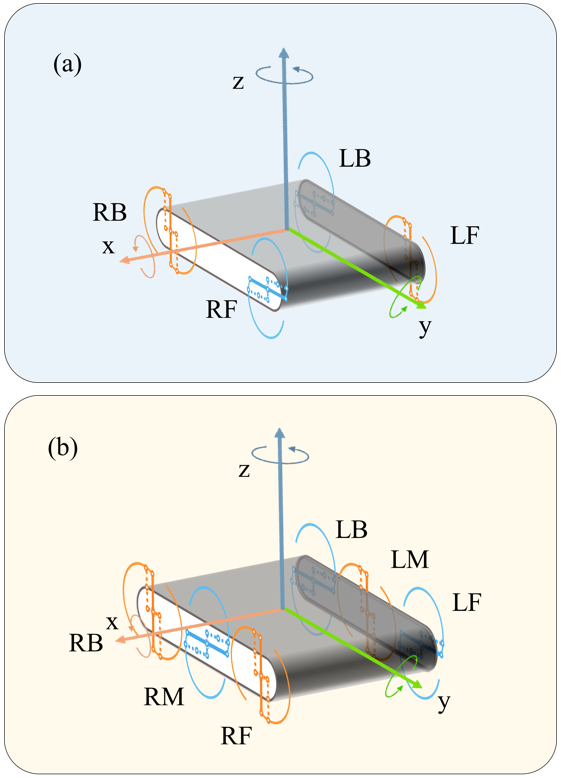}
    \vspace{-3mm}
    \caption{Leg and frame definition for (a) quadrupedal and (b) hexapod SWheg.}
    \label{fig:LegAndFrameDef}
    \centering
\end{figure}

Different gaits can be generated by applying phase difference $t_{k}$ to the wheel-legs. The following table shows the phase difference of some simple gaits and the corresponding gait diagram. This control strategy can also be easily transferred to robots with more wheel-leg modules. The definition of wheel-leg is shown in Figure~\ref{fig:LegAndFrameDef}. Figure~\ref{fig:gaitSch} shows the schematic of tripod gait and trotting gait.

\begin{table}[h!]
  \begin{center}
    \caption{Phase Diagram of Hexapod SWheg}
    \begin{tabular}{ccccccc} 
      \textbf{Gait Type} & \textbf{RF} & \textbf{RM}  & \textbf{RR}  & \textbf{LF} & \textbf{LM}  & \textbf{LR}\\
      \hline
      tripod & 0 & $\frac{1}{2} T_{c}$ & 0 & $\frac{1}{2} T_{c}$ & 0 & $\frac{1}{2} T_{c}$ \\
    \end{tabular}
  \end{center}
\end{table}

\begin{table}[h!]
  \begin{center}
    \caption{Phase Diagram of Quadrupedal SWheg}
    \begin{tabular}{ccccc} 
      \textbf{Gait Type} & \textbf{RF} & \textbf{RR} & \textbf{LF} & \textbf{LR}\\
      \hline
      Walk & 0 & $\frac{1}{4} T_{c}$ & $\frac{2}{4} T_{c}$ & $\frac{3}{4} T_{c}$ \\
      \hline
      Trot & 0 & $\frac{1}{2} T_{c}$ & $\frac{1}{2} T_{c}$ & 0 \\
    \end{tabular}
  \end{center}
\end{table}

\subsubsection{Turning in place}
The controller for in-place turning employs the same trajectories as in walking and trotting, except that the Legs on different sides rotate in opposite directions. The angular velocity of in-place turning mainly depends on the trajectory parameters $t_c$ and $t_{\delta}$.

\subsubsection{Synchronized Mode}
A naive control strategy was designed for stair climbing tasks presented in Section~\ref{stair experiment}. In this mode, the position of wheel-legs on the same axis (RF and LF, RM and LM, LB and RB) will be synchronized. Furthermore, to reduce the oscillation of the robot body, the Legs will rotate at a fixed velocity.


  
  
  
\section{Experiments} \label{experiments}
In this section, the tendon-drive wheel-leg transformable module, and its performance on different terrains are evaluated.

\subsection{Transformable Module Experiment}
\begin{figure*}[t!]
    \centering
    \includegraphics[width=5in]{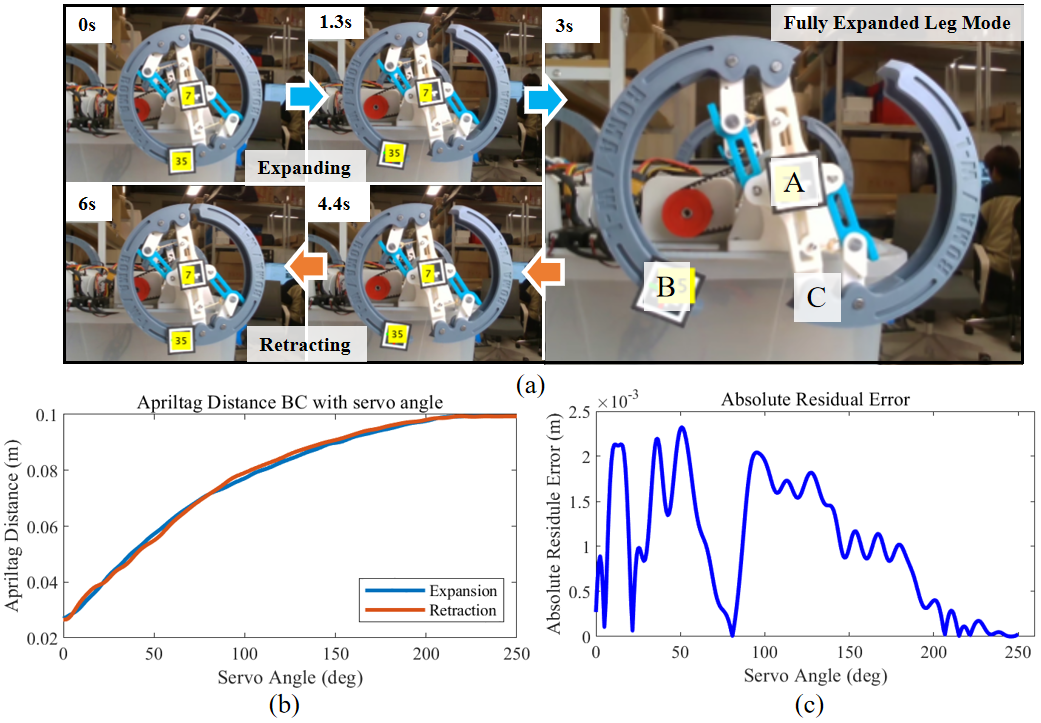}
    \vspace{-3mm}
    \caption{Transformation mapping experiment. (a) The whole transformation process measured: expanding and retracting. (b) The distance between two AprilTags $B$ and $C$ over the rotation angle of the servo. (c) The absolute residual error between the expanding and contracting process is also shown.}
    \label{fig:mapping}
    \centering
\end{figure*}

In this experiment, the accurate mapping between the servo's rotation angle and the transformation extent, represented by the distance between $BC$ as shown in Figure~\ref{fig:mapping}, is measured to ensure precise control over the transformation process. The rotational angle can be directly obtained from the built-in encoder of the servo. The distance $AB$ is identified by AprilTags. As shown in Figure~\ref{fig:mapping}a, we placed the AprilTag at joints $A$ and $B$, then used the calibrated camera to take pictures and calculate the spatial position $(x, y, z)$ of the two points $A$ and $B$ with respect to the camera coordinate system in the real world and calculate the distance between $AB$. Then we can solve for the distance between $BC$ via geometric relationships. We performed $10$ repetitive expanding and retracting experiments and recorded the corresponding data, the average mapping is utilized as the transformation standard and is shown in Figure~\ref{fig:mapping}b.

As shown in the absolute residual error plot given at the bottom right corner in Figure~\ref{fig:mapping}b, the maximum error is around $2mm$, $2\%$ of the maximum distance between $BC$, which validated the stability of the transformation process.
\subsection{Stair Ascending Experiment} \label{stair experiment}


To demonstrate the passability of the SWheg module on hard terrains, we tested the robot on two typical kinds of terrains: steps and stairs. The experiment is carried out in both simulation and real-world environments. 

\subsubsection{Simulation Setup}

We conducted simulation experiments on the Gazebo simulation platform. The imported parts are generated from the original CAD file. However, to boost the speed of simulation, the $4$-bar linkage close chain structure of the SWheg is replaced with a $2$-bar open chain structure in the simulation model. (see Figure~\ref{fig:simulation}) The removed parts are wrapped up by the rims and have no contact with the ground during operation. The tendon-driven transformation system is also neglected. The transformation is done by direct joint position control. Only the collisions of SWhegs are calculated. The simulation model uses one set of SWhegs modules only.

\subsubsection{Step Climbing Experiment}

In this experiment, we tested the maximum terrain height difference that SWheg can overcome when using three operation modes: Wheeled, legged, and reversed-legged. In legged mode, SWheg contacts the ground surface with its curve part. In reversed-legged mode, the tip part hits the ground. 

\begin{figure*}[t!]
    \centering
    \includegraphics[width=4.5in]{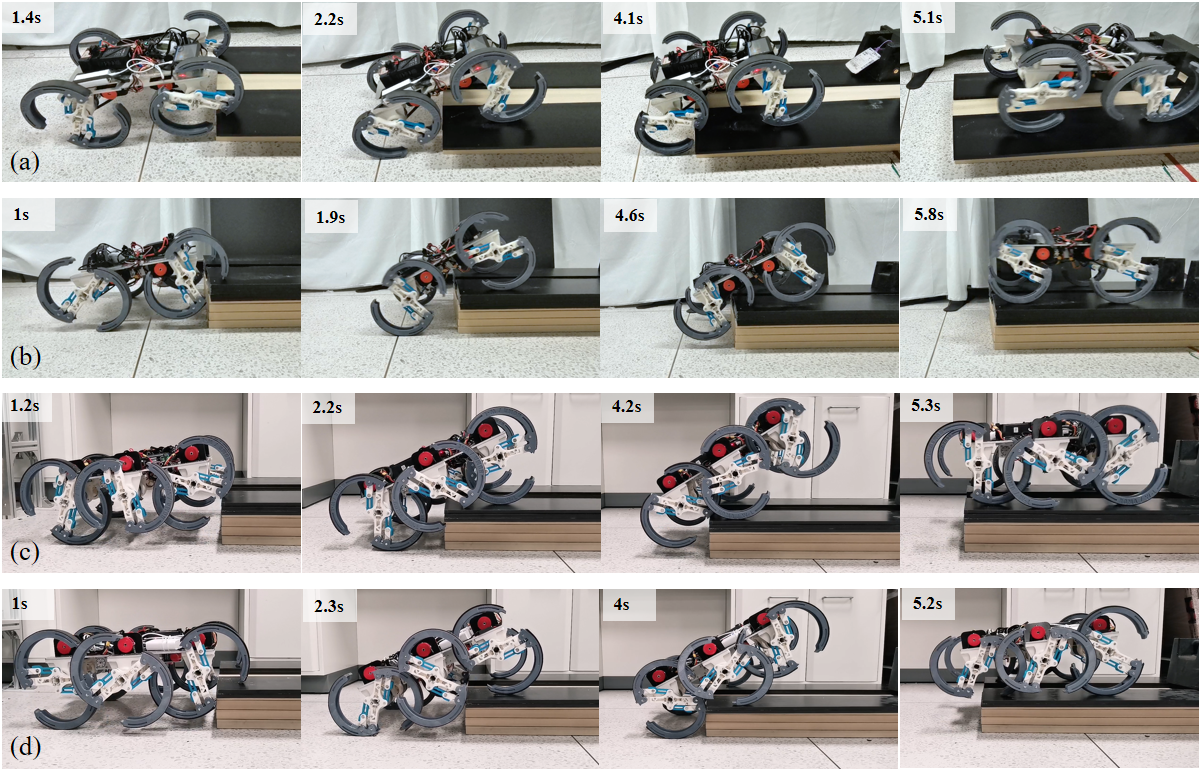}
    \vspace{-3mm}
    \caption{Pictures of SWheg's stair-climbing experiments. (a) and (b) Quadrupedal SWheg is climbing stairs in two directions.  (c) and (d) Hexapod SWheg is climbing stairs in two directions.}
    \label{fig:step}
    \centering
\end{figure*}

We performed experiments on steps of different heights and recorded the results. The pictures cut from the video are shown in Figure~\ref{fig:simulation}b. The test results of real robots (see Figure~\ref{fig:step}) are consistent with the simulation results. As shown in Table~\ref{tab:stepthreshold}, the performance of the legged modes is significantly better than the wheeled mode. 

\begin{table}[h!]
  \begin{center}
  \vspace{-3mm}
    \caption{Step Climbing Thresholds of Different Operation Mode}
    \begin{tabular}{ccccc} 
      \textbf{test mode} & \textbf{Wheel} & \textbf{Leg}  & \textbf{Leg(reversed)}  \\
      \hline
      simulation & 8 mm & 13 mm & 18 mm \\
      \hline
      real & 8 mm & 14.7 mm & 16.8 mm \\
      \label{tab:stepthreshold}
    \end{tabular}
  \end{center}
\end{table}

\subsubsection{Stair Climbing Experiment}

\begin{figure*}[t!]
    \centering
    \includegraphics[width=4.5in]{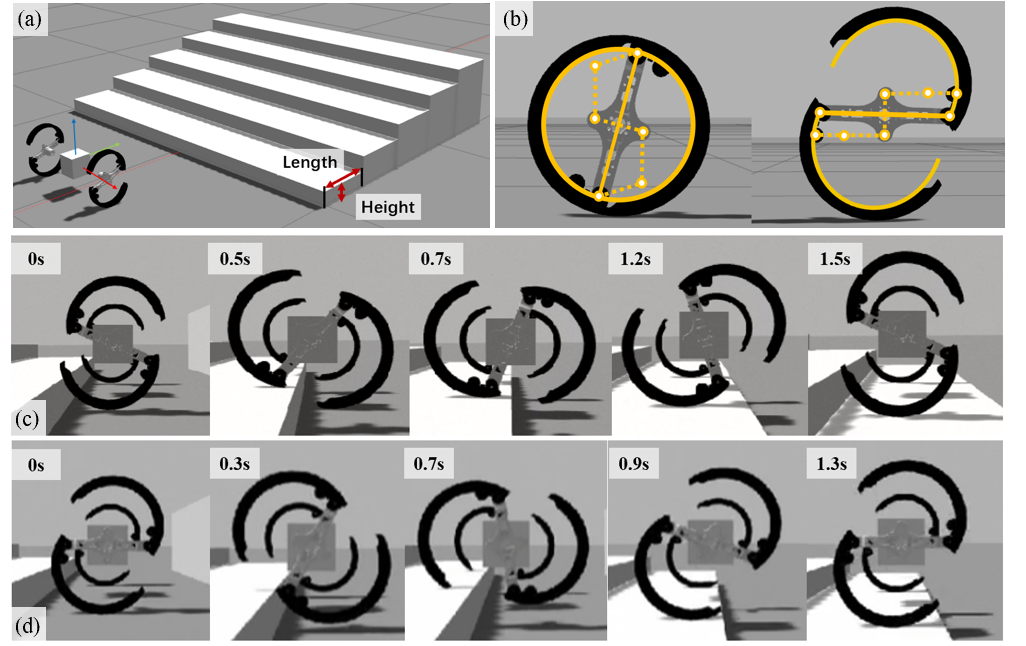}
    \vspace{-3mm}
    \caption{Pictures of simulation stair-climbing experiments. (a) The simulation test environment for the stair-ascending experiment. (b) Simplified SWheg wheel-leg module in simulation. The $4$-bar linkage is replaced with a 2-bar open chain structure. (c) Synchronized legged mode operation overcoming the obstacle. (d) Reversed legged mode operation overcoming the obstacle. }
    \label{fig:simulation}
    \centering
\end{figure*}

In this experiment, we tested the performance of SWhegs on continuous stairs. As shown in Figure~\ref{fig:simulation}, the two key parameters of stairs are stair depth and height. 

In the simulation, we generated $400$ test environments in total. The stair models are automatically generated with xacro and have the same physics configurations. As shown in Figure~\ref{fig:simulation}, the stairs models are fixed to the world. Each test environment contains five stairs. Both height and length of stairs range from $1mm$ to $20mm$. Additionally, a pose constraint is applied to the robots to ensure they face the stairs in the right direction.

\begin{figure}[t!]
    \centering
    \includegraphics[width=2.5in]{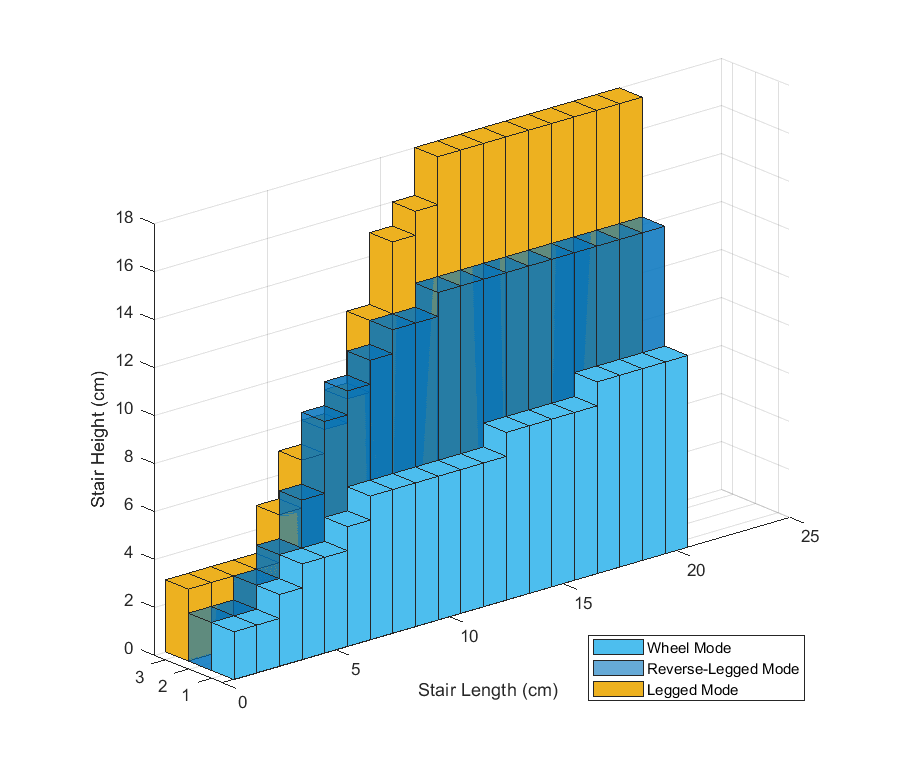}
    \vspace{-3mm}
    \caption{The height and depth of stair that SWheg can overcome in different modes. }
    \label{fig:SOZ}
    \centering
\end{figure}

The suggested operating zone (SOZ) is shown in Figure~\ref{fig:SOZ}. Stair length will be the major constraint when it is small. When the stairs have enough length, the situation will be more likely to be the 'step' situation. For stairs lower than $12mm$, the Legged mode has the best performance. For higher stairs, only the reverted-legged mode can get over the stairs.

\subsection{Motion Smoothness Experiment}

In this experiment, we measured the stability of SWheg on the ground using different gaits. A WITMotion-HWT906 IMU is installed on the chassis of the robot. The body frame is defined in Figure~\ref{fig:LegAndFrameDef}. The robot's orientation is represented by a set of $Z-Y-X$ Euler angles, $\phi,\theta,\psi$. Where $\psi$ is the yaw, $\theta$ is the pitch, and $\phi$ is the roll. When the robot is operating on flat ground, the pitch and roll angle is supposed to be 0. So, the stability of motion is evaluated with a cost function, $J(\phi,\theta,\psi)$.

\begin{equation}
J(\phi,\theta,\psi) = K_{s} [\theta^{2} + \phi^{2}]
\end{equation}

\begin{figure*}[t!]
    \centering
    \includegraphics[width=5.3in]{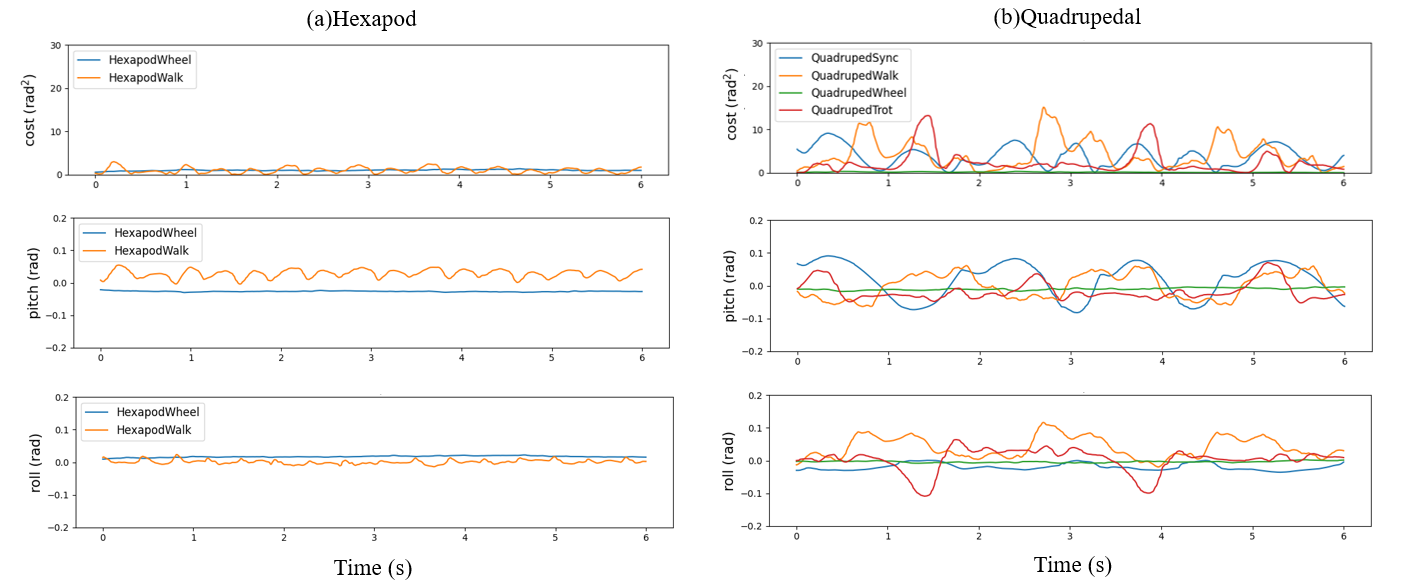}
    \vspace{-3mm}
    \caption{Motion smoothness test result of (a) hexapod and (b) quadrupedal SWheg robot. }
    \label{fig:motionSmoothness}
    \centering
\end{figure*}

Where $K_{s}$ is a constant scaling factor. Here we take $K_{s} = 1000$. Figure~\ref{fig:motionSmoothness} shows the motion stability of hexapod and quadrupedal SWheg in a $6$-seconds ($2$ gait cycles of legged modes) operation on flat ground. 

Both hexapod and quadruped SWheg have good stability when using wheeled mode. However, in legged modes, the tripod gait of the hexapod SWheg has better performance, which results from the internal stability of the hexapod configuration.

Performance of trotting and walking gait of quadrupedal SWheg significantly suffers due to the lack of support. Most quadruped wheel-leg transformable robots \cite{chen2017turboquad,chen2013quattroped,chen2014Quattroped} use two or more actuators on each leg to acquire better stability. However, since we are using only one motor for each SWheg, the system is highly underactuated and unstable. 

To achieve better stability on quadruped SWheg, combining walking mode with synchronized mode (see 4.3.1) is recommended. In synchronized mode, all four SWhegs have reliable contact with the ground. However, the synchronized mode lacks turning ability and must be used together with other walking modes.

\subsection{Power Efficiency}
\subsubsection{Speed and Specific Resistance}
Wheeled mode and legged mode have their locomotion characteristics on different terrains. Wheeled mode contact the ground continuously and has high movement performance on even terrain, while legged mode can pass on uneven terrains with its discontinuous characteristic. In this study, we combined the advantages of wheeled mode and legged mode by integrating both modes on a single platform with minimal actuator realization.


In previously studies on wheel-leg transformable robots, the power efficiency performance of the robots was evaluated using specific resistance(SR)\cite{RN545}. For instance, Turboquad and Quattroped gave detailed SR performance of the platform on different surfaces (i.e. linoleum, asphalt, and grass) with different forward speeds.

For intuitive comparison, we evaluate the performance of the performance using SR as a standard. The experiment was conducted 
on each mode of quadrupedal and hexapod robot separately.
\begin{equation}
    SR=\frac{P}{mgv}
\end{equation}
Where $m$ is the total weight of the robot, where the mass of quadrupedal SWheg and the hexapod SWheg is shown in Table~\ref{tab:ROBOT SPECIFICATIONS}. $v$ is the averaged forward speed, and $P$ is the averaged power consumption. The power consumption measured here is the total power consumed directly detected by a power measurement module. And this power consumption contains the power from all the actuators but not from the sensors and the processor.

\begin{figure}[t!]
    \centering
    \includegraphics[width=2.5in]{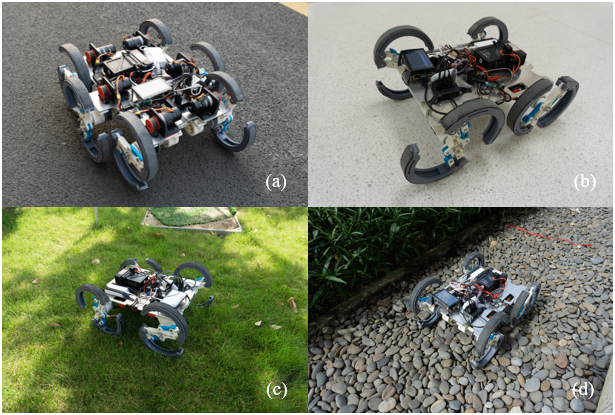}
    \vspace{-3mm}
    \caption{Power efficiency evaluation experiments on various terrains including (a) asphalt, (b) flat ground, (c) grass and (d) pebbles.}
    \label{fig:outdoor}
    \centering
\end{figure}

\begin{figure*}[!p]
    \centering
    \subfloat[Quadrupedal SWheg data.\centering ]{\includegraphics[width=4.7in]{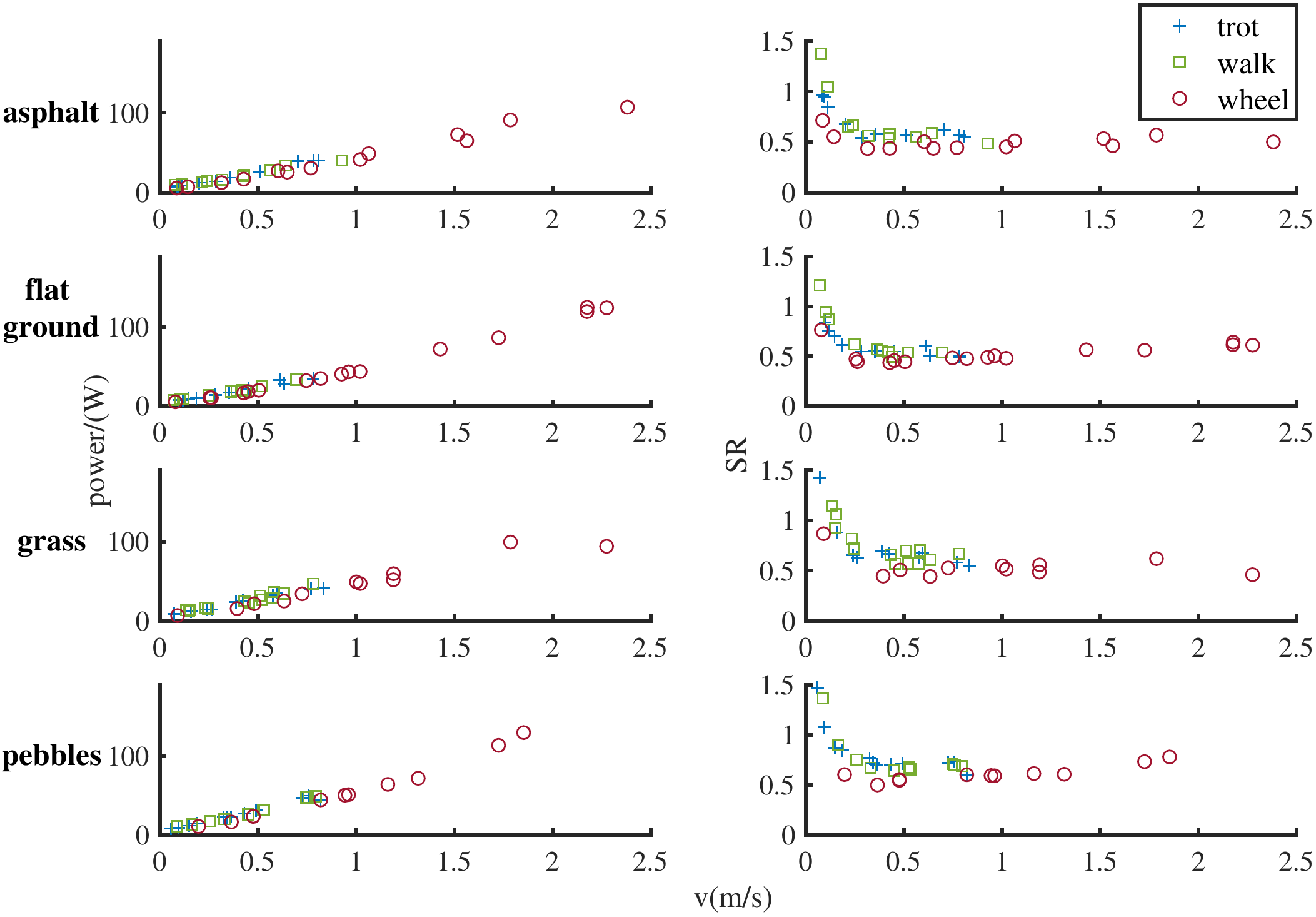} \label{sr_Quadruped}}
    \vfill
    \subfloat[Hexapod SWheg data.\centering ]{\includegraphics[width=4.7in]{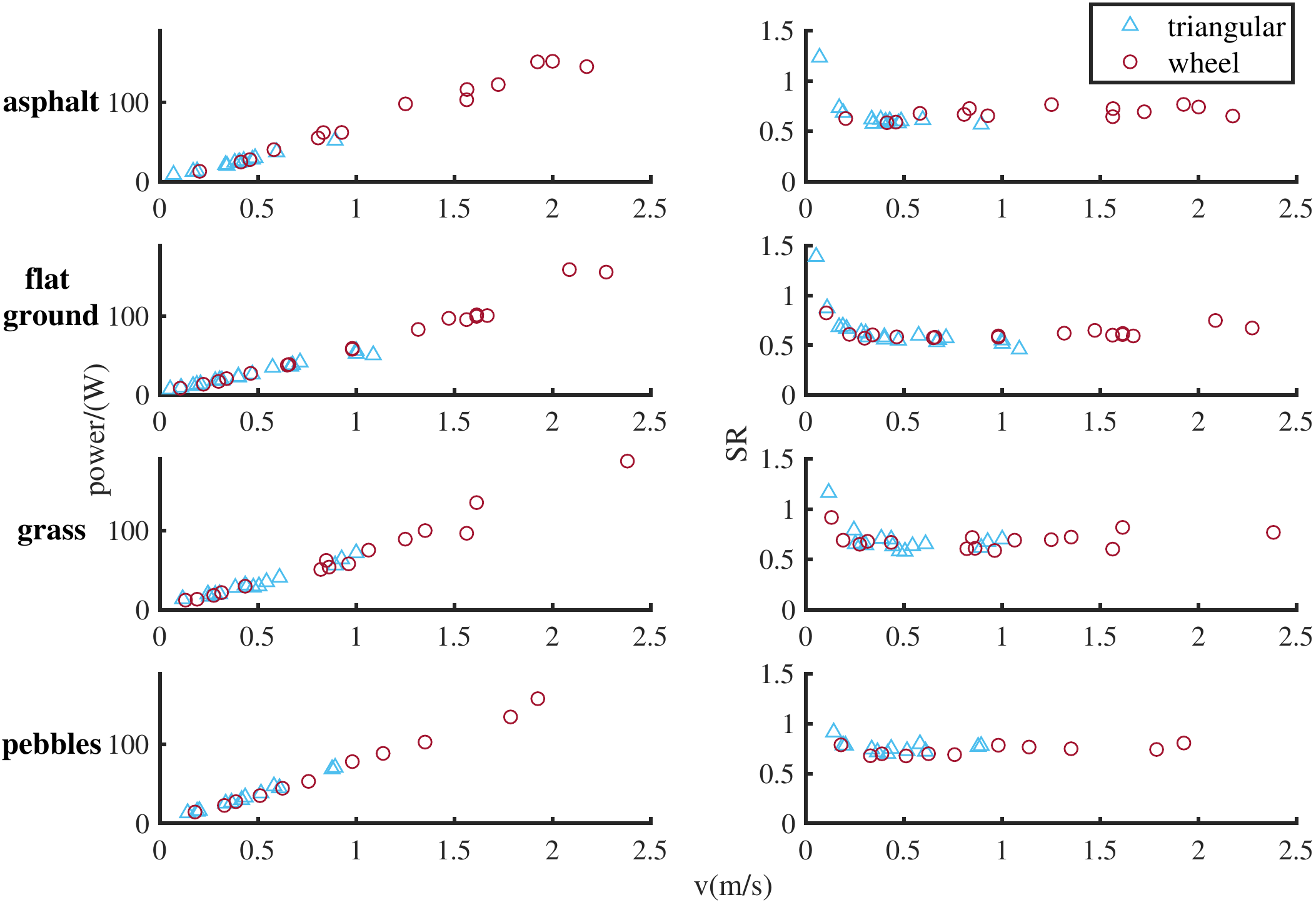} \label{sr_Hexapod}}
    \caption{Power and Specific Resistances of (a) Quadrupedal and (b) Hexapod while running in four terrains with different gaits/modes.}
    \label{fig:fig_powerSR}
\end{figure*}

In this experiment, we tested two robots on different terrains(flat ground, asphalt, grass, pebbles) with legged mode and wheeled mode, which is shown in the Figure~\ref{fig:outdoor}a,d, where three different gaits/modes for quadrupedal SWheg, including wheeled mode, trotting gait, and walking gait, two gaits/modes for hexapod SWheg, including wheeled mode and triangular gait. We recorded several powers and running times for each gait/mode and computed the forward velocity and specific resistance to plot the result on Figure~\ref{fig:fig_powerSR}.


Overall power consumption is mainly determined by the behavior of motor operation. Figure~\ref{fig:fig_powerSR} shows SR versus the forward speeds for two robots. Several conclusions can be derived as follows:

(1) For all forward speeds of the quadrupedal SWheg robot in four different terrains, the wheeled mode is more power efficient than other gaits, including walking gait and trotting gait. This result corresponds to our expectation, in the legged mode of quadrupedal SWheg, the robot need to consume more energy for the ground contact energy loss and the leg acceleration and deceleration which is based on position-based trajectory planning. But for the hexapod SWheg robot, the SR values are very similar in four different terrains for two locomotion modes, where the result is very different from that of the quadrupedal. The most likely reason is that when the hexapod Swheg robot moves in the triangular gait, there are always three legs touching the ground, in other words, due to the smoother motion of the triangular gait compared with other gaits, it minimizes the contact energy loss considerably.

(2) All the measurements in legged mode, i.e. trotting gait, walking gait, and triangular gait, were acquired in the speed range of about $0\sim1m/s$, yet the wheeled mode measurements were recorded in the speed range of $0\sim2.5m/s$. The reason is that the legged mode is using position-based trajectory planning and has an acceleration and deceleration process, and there is a big impact when the acceleration happened the ground contact, which increases the risk of robot damage. Despite this speed limit for legged mode, it still has better performance when encountering rough terrains with various obstacles.  Based on this finding, we can land on the mode strategy for the SWheg robots: when moving in these four different terrains in the speed range of  $1\sim2m/s$, use wheeled mode. When moving in rough terrains with obstacles, transforms into legged mode.

(3) The overall power consumption of the robot increases with the forward speed, but the specific resistance, i.e. cost of transport, decreases with faster speed.  The power efficiency trend in four different terrains of quadrupedal and hexapod SWheg robots looks very similar. All of the SR values decrease evidently in the range of $0\sim0.5m/s$ and after that tend to steadily.  This result indicates that the robot is more energy-efficient in high-speed motion. The reason is that the SR value is a time-relevant value, when a robot moves slowly in the terrain, it consumes more energy on motor heat loss and so decreases the power efficiency.

(4) The type of terrain has a huge effect on power consumption. Both quadrupedal and hexapod SWheg robots are more power efficient on flat ground and asphalt compared with grass and pebble terrain, as the locomotion in rough terrains consumes more energy.

(5) In wheeled mode, the SR value of a hexapod SWheg robot is higher than that of the quadrupedal SWheg robot when moving in these four different terrains. One of the possible reasons is that when moving at the same speed in wheeled mode, the Hexapod SWheg robot has two more wheels in contact with the ground, and the corresponding friction force increases the cost of transport. In the legged mode, including the triangular gait of the hexapod SWheg robot, and the walking and trotting gait of a quadrupedal SWheg robot, the SR value is very much similar. This indicates that these gaits are equivalent power-efficiency-wise, and can be selected based on other criteria such as forward speed or motion smoothness.

\subsubsection{Energy Consumption on long terrains}
To validate SWhegs' advantage of seamlessly integrating wheels and legs on a single platform, we conducted comparison tests on a long terrain consisting of various terrains using three different locomotion strategies, namely: only-wheel, only-leg, and wheel-leg integrated strategies. 
The tests were performed outdoors on a terrain containing grass, pebbles, flat ground, and stairs in the route. The only-wheel strategy fails at steps greater than the radius of the wheel. The only-leg strategy successfully finished the route, where the Quadruped robot used $65.6s$ with $896.0J$ energy consumption in total, hexapod SWheg robot used $48.9s$ with $1029.5J$ energy consumption in total. In contrast, the wheel-leg integrated strategy took $41.3s$ with $666.1J$ energy consumption for quadruped and $37.3s$ with $995.7J$ energy consumption for hexapod. To sum up, this experiment validates the overall better performance of wheel-leg integrated platforms, which is the ability to overcome obstacles higher than wheel radius compared to only-wheel platforms and is faster and more energy efficient than only-leg platforms.
\section{Conclusion and Future Work} \label{conclusion}

This paper reported the design and performance evaluation of a novel modular wheel-leg transformable robot design with minimal actuator realization named SWheg. The robot can perform fast morphology transformation between wheels and S-shape legs using only $50\%$ actuators when compared to the literature designs. This greatly reduces system complexity, cost, and failure rate. The proposed design has a novel wheel-leg transformable module that uses tendon-driven motion mode switching and a tendon network that uses a single actuator to switch the motion mode of the entire robot. We built two platforms with four and six wheel-leg modules to validate our design principle and compared the performance of the platforms in both simulation and various terrains. Our robot shows great overall mobility and obstacle negotiation capability. In the future, we will further improve the design to make the system more reliable and easy to configure. We will also develop specialized behaviors to enable online gait transitions and faster movements. On the perception side, We will add sensing techniques to the platform such as depth camera and radar to enable the robot to classify and negotiate with different terrains by adjusting gaits and modes.

\section*{Acknowledgments}
This work was supported in part by the Science, Technology and Innovation Commission of Shenzhen Municipality under grant no. ZDSYS20200811143601004.

\printendnotes

\bibliographystyle{apalike}
\bibliography{main.bib}

\end{document}